\documentclass[10pt,letterpaper]{article}
\usepackage[top=0.85in,left=2.75in,footskip=0.75in]{geometry}

\usepackage{amsmath,amssymb}

\usepackage{changepage}

\usepackage{textcomp,marvosym}

\usepackage{cite}

\usepackage{nameref,hyperref}

\usepackage{multirow}

\usepackage[nopatch=eqnum]{microtype}
\DisableLigatures[f]{encoding = *, family = * }

\usepackage[table]{xcolor}

\usepackage{array}

\newcolumntype{+}{!{\vrule width 2pt}}

\newlength\savedwidth

\newcommand\thickhline{\noalign{\global\savedwidth\arrayrulewidth\global\arrayrulewidth 2pt}%
\hline
\noalign{\global\arrayrulewidth\savedwidth}}

\raggedright
\usepackage[aboveskip=1pt,labelfont=bf,labelsep=period,justification=raggedright,singlelinecheck=off]{caption}

\newcounter{quotecount}
\newcommand{\MyQuote}[1]{\vspace{0.2cm}\addtocounter{quotecount}{1}%
     \parbox{11.5cm}{\em #1}\hspace*{0.1cm}(\arabic{quotecount})\\[0.3cm]}
     
\makeatletter
\renewcommand{\@biblabel}[1]{\quad#1.}
\makeatother

\usepackage{lastpage,fancyhdr,graphicx}
\usepackage{epstopdf}
\fancyheadoffset[L]{2.25in}
\fancyfootoffset[L]{2.25in}
\begin{document}

\epstopdfDeclareGraphicsRule{.tif}{png}{.png}{convert #1 \OutputFile}
\AppendGraphicsExtensions{.tif}

\vspace*{0.2in}

\begin{flushleft}
{\Large
\textbf\newline{Unveiling factors influencing judgment variation in Sentiment Analysis 
with Natural Language Processing and Statistics} 
}
\newline

Olga Kellert \textsuperscript{1*},
Carlos Gómez-Rodríguez \textsuperscript{1},
Mahmud Uz Zaman \textsuperscript{2}


\bigskip
\textbf{1} Universidade da Coruña, CITIC, Grupo LyS, Depto. de Ciencias de la Computación y Tecnologías de la Información, Campus de Elviña s/n, 15071 A Coruña, Spain
\\
\textbf{2} University of Augsburg, The Applied Computational Linguistics (ACoLi) Lab, Universitätsstr. 10, 86159 Augsburg, Germany


\bigskip

%
%
* OK has the first authorship on this work.



\textpilcrow Acknowledgements: European Research Council (ERC), which has funded this research under the Horizon Europe research and innovation programme (SALSA, grant agreement No 101100615), MICIU/AEI/10.13039/501100011033 (SCANNER-UDC, PID2020-113230RB-C21), Xunta de Galicia (ED431C 2020/11), and Centro de Investigación de Galicia “CITIC”, funded by Xunta de Galicia and the European Union (ERDF - Galicia 2014–2020 Program), by grant ED431G 2019/01.

\textpilcrow Comments: Accepted manuscript to be published in PLoS One.

* Corresponding author: kellert-olga@gmx.de

\end{flushleft}
\section*{Abstract}
TripAdvisor reviews and comparable data sources play an important role in many tasks in Natural Language Processing (NLP), providing a data basis for the identification and classification of subjective judgments, such as hotel or restaurant reviews, into positive or negative polarities. This study explores three important factors influencing variation in crowdsourced polarity judgments, focusing on TripAdvisor reviews in Spanish. Three hypotheses are tested: the role of Part Of Speech (POS), the impact of sentiment words such as ``tasty'', and the influence of neutral words like ``ok'' on judgment variation. The study’s methodology employs one-word titles, demonstrating their efficacy in studying polarity variation of words. Statistical tests on mean equality are performed on word groups of our interest. The results of this study reveal that adjectives in one-word titles tend to result in lower judgment variation compared to other word types or POS. Sentiment words contribute to lower judgment variation as well, emphasizing the significance of sentiment words in research on polarity judgments, and neutral words are associated with higher judgment variation as expected. However, these effects cannot be always reproduced in longer titles, which suggests that longer titles do not represent the best data source for testing the ambiguity of single words due to the influence on word polarity by other words like negation in longer titles.  This empirical investigation contributes valuable insights into the factors influencing polarity variation of words, providing a foundation for NLP practitioners that aim to capture and predict polarity judgments in Spanish and for researchers that aim to understand factors influencing judgment variation.



\section*{Introduction}
Sentiment Analysis (SA) holds a pivotal role in Natural Language Processing (NLP), dealing with the identification of subjective assessments, such as opinions on hotels or restaurants in textual content (subjectivity identification task), and predicting the polarity of these assessments, i.e. whether they express a positive or negative view (Polarity Classification task, PC), as exemplified in (1) \cite{cui2023survey}. 

\MyQuote
{\textbf{Anne}: This hotel is awesome. (PC: Positive )

\textbf{Ben}: This hotel is terrible. (PC: Negative)
}

While all SA approaches in NLP share the fundamental idea that subjectivity and polarity are encoded in natural language to make a prediction about subjectivity and polarity even possible, they differ in their assumptions regarding whether single (lexical) words play a central role in this prediction. Some approaches, known as dictionary-based methods, rely on specific sentiment words to perform SA tasks like PC\cite{cui2023survey,vilares2017universal,kellert2023experimenting,hutto2014vader}. In examples like (1), we, humans, intuitively attribute subjectivity and polarity to words that describe the evaluated entity or target, such as ``awesome'' and ``terrible'', also referred to as subjective, evaluative predicates or Personal Taste Predicates (PTPs) in Linguistics\cite{lasersohn2005context,kennedy2013two,umbach2020evaluative,kennedy2022familiarity} and as sentiment words in Sentiment Analysis and Opinion Mining\cite{cui2023survey,vilares2017universal,kellert2023experimenting,hutto2014vader}. This hypothesis, namely that words can encode subjectivity and polarity lexically, resonates with linguistic and philosophical discussions\cite{lasersohn2005context,kennedy2013two,umbach2020evaluative,kennedy2022familiarity}.  However, the question of which words encode subjectivity and polarity lexically in virtue of their meanings and which words require further contextual information or information about the communication context to express subjectivity and/or polarity is a matter of ongoing debate \cite{lasersohn2005context,kennedy2013two,umbach2020evaluative,kennedy2022familiarity}. Consider the subjective statement in (2). 

\MyQuote
{This hotel is ok. (PC: ?)}

The word ``ok'' likely does not possess a strongly positive polarity like ``awesome'' in (1) and may be interpreted positively or negatively by different individuals, as in ``this hotel is ok, but nothing special.''  or as in ``this hotel is ok, it’s actually better than I expected.'' However, even if some people tend to interpret the word ``ok'' more positively and some people more negatively, it is very unlikely that the polarity of ``ok'' has the same strength as the polarity of ``awesome'' and ``terrible'' in (1) by the virtue of the lexical meaning of ``ok''. We thus expect to find some polarity variation of words like ``ok'' that can be described as neutral in contrast to positive or negative words, but this variation will be restricted by the lexical meaning of this word. The difference between words like (1) and (2) has been observed in the linguistic literature \cite{kellert2021free,alonso2021random} and has been indirectly stated in NLP\cite{alvarez2023overview}. Alvarez-Carmona et al. (2023) have observed that all NLP approaches that participated in the Shared Task Rest-Mex 2023 with the goal of classifying automatically the polarity of TripAdvisor Reviews in Mexican Spanish produced considerably more errors in the classification of reviews with neutral polarity than in the classification of reviews with extremely positive or extremely negative polarity. The error analysis could be seen as an indirect confirmation of the difference between words in (1) and (2). However, this hypothesis has not been tested on polarity variation of words yet. 

A different case is the predicate ``on the highway'' in (3), which does not seem to encode polarity lexically at all, as the statement in (3) can be perceived as (strongly) positive by those who prefer proximity to the highway or as (strongly) negative by those who do not favor proximity to the highway or as neutral by those who just do not care about the location of the hotel at all. In Linguistics, expressions like ``on the highway'' in (3) would not even be evaluated as subjective as the sentence in (3) is either objectively false or true \cite{lasersohn2005context,kennedy2013two,umbach2020evaluative,kennedy2022familiarity}. In short, the expression “on the highway” will probably show a full range of polarity variation that depends on meaning external factors like individual preferences. This prediction means that people will much more strongly disagree in judgments using expressions like “on the highway” than using sentiment words like in (1).  

\MyQuote
{This hotel is on the highway. (PC: ?)}

The question then arises whether we can distinguish between the three cases in (1), (2) and (3) quantitatively on the basis of word polarity variation. The main assumption is that the positive and negative PTPs like in (1) have the smallest polarity variation, that is, people who use these words have the smallest judgment disagreement. Neutral words like in (2) have a higher polarity variation than words in (1) and words like in (3) have a full polarity variation. We assume an order of polarity variation going from the highest polarity variation to the smallest variation from left to right:

\MyQuote
{Non-subjective predicates (e.g. ``on the highway'') 

\textbf{\textgreater} PTPs referring to middle scale (e.g. ``ok'') 

\textbf{\textgreater} PTPs referring to extreme ends of a scale (e.g. ``terrible'')}

Another factor we investigate in this article is the influence of the Part of Speech (POS) of content words such as adjectives, nouns, verbs and adverbs on polarity variation. The POS of content words has been studied in the research on polarity ambiguity of words in languages like Chinese and English in the SemEval 2010 Task 18 \cite{wu-jin-2010-semeval,Xia}. In our preliminary study, we examined the difference between nouns and adjectives in Spanish \cite{Kellert}. However, a complete study of all POS types of content words and their relation to polarity variation has not been studied for Spanish yet. 

The factors influencing polarity variation of Spanish words remain largely unexplored in NLP, as well as in Linguistics and Philosophy (see the Related Work section for details). Prevalent NLP approaches for SA are agnostic about the role of sentiment words or lexical words in general. Instead, they utilize the entire input text, harnessing Machine Learning and Deep Learning techniques \cite{devlin-etal-2019-bert,mikolov2013efficient} for tasks like polarity classification \cite{alvarez2023overview},\cite{cui2023survey}. While powerful, these approaches lack transparency in explaining how they leverage linguistic features such as sentiment words for prediction. However, recent studies that combine sentiment dictionaries with Deep Learning methods, referred to as hybrid approaches, have shown superior performance in SA tasks (\cite{cui2023survey}). This underscores the importance of investigating the polarity variation of individual words, not only for evaluating dictionary-based approaches, but also for enhancing hybrid approaches. In this paper, we undertake this study. The results of this research will furnish NLP practitioners, regardless of their approach (dictionary-based or hybrid), with an empirical foundation for subjectivity and polarity classification, essential for effective Sentiment Analysis (SA). Furthermore, the results of this study will contribute to the theoretical debate in Linguistics and Philosophy about the extent to which words inherently encode polarity. The results will supply empirical insights into the theoretical debate. 


The remainder of the paper is structured as follows: the next section presents related work in more detail, as well as our research hypothesis. This is followed by a Materials and Methods section, outlining the datasets and methodology for hypothesis testing. Then, we have a section to present and discuss the results; and a final section for the conclusion and outlook of the paper.

\section*{Related Work and Research Hypotheses}

\label{sec:relatedwork}

Previous work on subjectivity and polarity in Linguistics is centered on predicative adjectives or nouns as in ``Chocolate is tasty'' or ``Jumping is fun'' \cite{lasersohn2005context,kennedy2013two,umbach2020evaluative,kennedy2022familiarity} and their semantic classification. Predicative nouns and adjectives are words that are arguments of predicative verbs like ``be'', ``seem'', etc. as in ``Chocolate is tasty''. Despite the numerous studies on lexical polarity in Linguistics that focus on the polarity of indefinites like ``some'' or ``none'' (see, \cite{israel1996polarity} for an overview), there are no quantitative studies on lexical polarity of sentiment words similar to ``tasty'' and ``fun'' in other syntactic contexts than predicative words \cite{lasersohn2005context,kennedy2013two,umbach2020evaluative,kennedy2022familiarity}. This lack in quantitative studies is partially explainable by the general trend in Linguistics using experts’ knowledge and/or intuitions in linguistic studies on subjectivity and polarity \cite{lasersohn2005context,kennedy2013two,umbach2020evaluative,kennedy2022familiarity}. More recently, subjectivity and polarity have been also investigated quantitatively in NLP (see\cite{cui2023survey} for an overview of NLP approaches on subjectivity and polarity). However, mainstream NLP approaches do not address the question of lexical polarity and subjectivity \cite{cui2023survey} and the only approaches that indirectly address this question are dictionary-based approaches that use sentiment words, which presumably encode subjectivity and polarity lexically \cite{brooke-etal-2009-cross,baccianella-etal-2010-sentiwordnet}. However, sentiment dictionaries are often built on manually selected sentiment words by few experts \cite{brooke-etal-2009-cross}. It is thus unclear whether sentiment words from sentiment dictionaries of Spanish contain an exhaustive list of sentiment words with lexical polarity or with a polarity that is inherent to these words. In our preliminary study, we have evaluated the sentiment dictionary of Spanish SO-CAL \cite{brooke-etal-2009-cross} showing that the sentiment words used in this dictionary are not optimally chosen as they are statistically more ambiguous than non-manually chosen words from one-word titles and other datasets \cite{Kellert}. The influence of POS of content words like verbs and adjectives on word polarity is not well explored yet either. There are few studies on this topic \cite{wu-jin-2010-semeval,Xia}, \cite{cao2015sentiment,wang2015fuzzy,Yin2020TheCO}. However, these studies are based on other languages than Spanish and most of these studies do not contain statistical information on ambiguity measures. While the dataset of\cite{wu-jin-2010-semeval} includes manual annotations of ambiguous adjectives in Chinese and English, the authors in\cite{Xia} also consider other word classes than just adjectives in the ambiguity study, showing that adjectives and verbs make up the most frequent POS of ambiguous words in their study. However, these results are based on few languages and it is thus not clear whether the same effects exist in Spanish. Moreover, previous studies on POS and polarity ambiguity of words do not investigate the relation of POS and other variables such as sentence length. In our preliminary study, we have shown that sentence length plays an important role in lexical ambiguity \cite{Kellert}. It is thus important to consider the sentence length in the study of the relation between word ambiguity and POS. We have also shown in our preliminary study that the distinction between content words like adjectives and functional words or stop words like determiners as well as the difference between nouns and adjectives is statistically important for ambiguity \cite{Kellert}. In this study, we aim to test other word types of content words including verbs and adverbs and their relation to ambiguity. In addition, we want to test whether the semantic class (+/- sentiment words or +/-PTPs) such as sentiment words in (1) and non-sentiment words in (3) plays a role for ambiguity. Finally, we test whether the polarity category (neutral, positive or negative) matters for ambiguity assuming that neutral words like in (2) will be more ambiguous than positive or negative words as in (1). 

Our first hypothesis is that the POS of content words matters for ambiguity (Hypothesis 1). Our second hypothesis is that the semantic class also matters for polarity distribution. Linguists assume that words that refer to personal experience of taste, smell, etc. like ``tasty'' (PTPs) encode subjectivity lexically \cite{lasersohn2005context,kennedy2013two,umbach2020evaluative,kennedy2022familiarity}, whereas words or statements that are objectively measurable like in (3) are not subjective. The subjective interpretation in (3) is not related to the word itself, but to the personal preference of the truth-value in (3). Our next hypothesis is thus that PTPs like ``tasty'' and ``fun'' influence polarity variation (Hypothesis 2). To test this hypothesis, we need to define the semantic class of words like Personal Taste Predicates and objectively measurable predicates, which is not a simple task and has not been undertaken so far in Sentiment Analysis in NLP. In fact, approaches for SA in NLP do not make any distinction between +/- objectively measurable predicates or predicates that can trigger false or true statements \cite{cui2023survey,alvarez2023overview, hill-korhonen-2014-learning,wiebe2000learning,choi2017coarse}. There are studies in NLP that focus on the identification of claims defined as “assertions about the world that can be checked” \cite{DBLP:journals/corr/abs-1809-08193}. However, these studies have not been related to the research on subjectivity and polarity. As a result, the distinction between +/- objectively measurable predicates has not been tested yet with NLP and corpus-linguistic methods in the domain of Sentiment Analysis. As developing NLP tools for the automatic distinction of +/- objectively measurable predicates would necessitate a whole project by itself, we use a manual annotation method of a relatively small dataset of ambiguous and non-ambiguous words to test the semantic class in combination with other heuristic methods (see § Materials and Methods). 

Based on the discussion of the examples in (1) and (2), we also assume that the polarity ambiguity is dependent on the polarity category (neutral, positive or negative) as neutral words can be interpreted either positively neutral or negatively neutral. The author in \cite{kellert2021free} has shown that words that belong to the middle polarity like ``normal'', ``ok'', etc. are often interpreted negatively in a certain word context as in ``He is (just) a normal guy. Nothing special.'' In some Romance varieties and languages including Spanish, words with middle or neutral polarity can change into negative polarity words \cite{kellert2021free}. For instance, the word ``vulgaire'' in Old French had previously the neutral meaning of ``common'' or ``normal'' and then changed to the negative meaning of ``rude'' or ``vulgar'' \cite{kellert2021free}. In short, we have reasons to believe that neutral words like ``ok'' are more ambiguous than words that refer to extreme polarities. Our next hypothesis is thus that words with a more neutral interpretation tend to be more ambiguous than words with positive and negative polarity (Hypothesis 3).
To summarize our hypotheses in (5).

\MyQuote
{\begin{itemize}
\item {Hypothesis 1: POS (adjectives, nouns, adverbs and verbs) influence polarity variation of words.}
\item {Hypothesis 2: Personal Taste Predicates influence polarity variation.}
\item {Hypothesis 3: Words with neutral polarity like ``ok'' are more ambiguous than words with positive or negative polarity like ``excellent'' and ``terrible''.}
\end{itemize}}

\section*{Materials and Methods}
\subsection*{Data}

For testing our hypotheses from section 2, we are using the training dataset of Rest-Mex 2023 from TripAdvisor, where Spanish-speaking tourists provide their evaluation or judgment on hotels, sightseeing’s and restaurants on a scale from 1 to 5, where 5 is the best evaluation \cite{alvarez2023overview}. The Rest-Mex dataset collection was gathered from various tourist destinations in Mexico, Cuba, and Colombia \cite{alvarez2023overview}. The data includes labeled information about polarity, type of attraction, and the country of origin for each opinion (see Table~\ref{table1}). This collection was obtained from the tourists who shared their opinion on TripAdvisor between 2002 and 2023. Each opinion’s polarity is an integer between [1, 5], where {1: Very bad, 2: Bad, 3: Neutral, 4: Good,5: Very good}. The collection includes 70/30 partition of the train and test dataset. For our study, we use the train dataset, which contains 251,702 labeled instances or reviews.

\begin{table}[!ht]
\centering
\caption{{\bf Example 4 from the Rest-Mex 2023 train dataset \cite{alvarez2023overview}}}
\begin{tabular}{|p{83pt}|p{60pt}|p{35pt}|p{37pt}|p{42pt}|}
\hline
{\bf Review}&
{\bf Title}&
{\bf Polarity}&
{\bf Country}&
{\bf Type}\\
\thickhline
Justo lo que buscaba. Sabores exoticos, buena atención, lugar tranquilo y bonito. Full recomendado. El sector también es tranquilo. &
Recomendado!&5&Colombia&Restaurant\\ \hline
‘Just what I looked for. Exotic flavors, good attention, quiet and nice location. Full recommendation. The area is also very quiet.’&Recommended!&&&\\ \hline
\end{tabular}
\label{table1}
\end{table}

\begin{figure}[!h]
\caption{{\bf Polarity Distribution in TripAdvisor reviews in Rest-Mex 2023 train dataset.}
The X-axis represents the polarity judgment on TripAdvisor reviews (from 1=worst to 5=best) provided by TripAdvisor users. The Y-axis represents the frequency of polarity judgments. }
\includegraphics[]{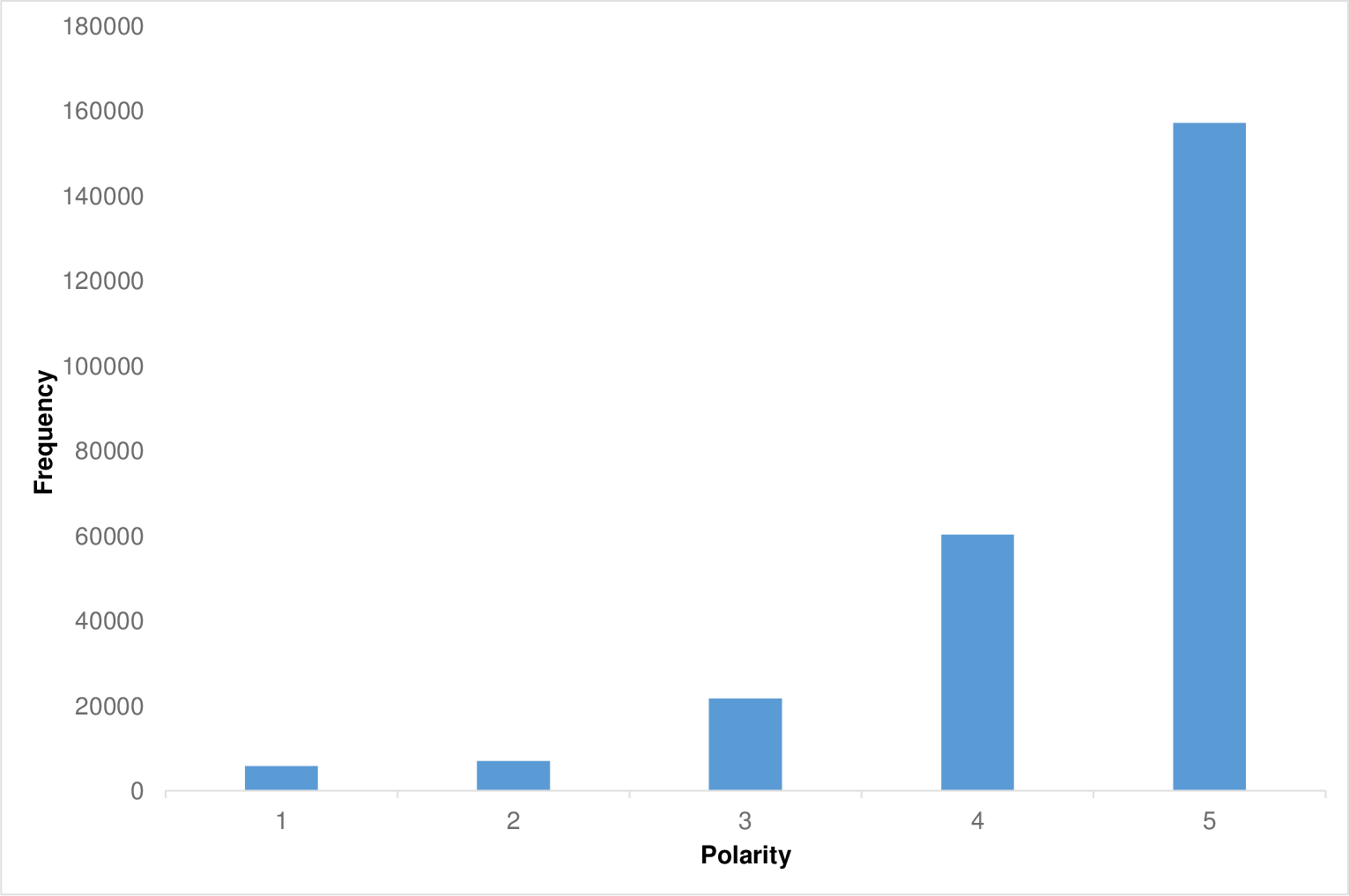}
\label{fig1}
\end{figure}

One important note about the dataset is that it is not balanced \cite{alvarez2023overview}. Polarity 5 is the most frequent polarity, followed by 4, 3, 2, and 1 as shown in Fig~\ref{fig1}. This bias in the data needs to be considered when evaluating the polarity distribution of words.

Using whole reviews as a basis for investigation of the polarity variation of single words is very challenging because words in full reviews appear in large contexts (up to 20 sentences) and the context can change or shift the polarity of the word such as negation as in ``not good''. In addition, long reviews contain various sentences that can be more representative of the final polarity judgment of the reviewer than other sentences \cite{kellert2023experimenting}. Figuring out what is the most important sentence for the polarity judgment adds noise to using full reviews for the study of lexical subjectivity and polarity. For these reasons, our basic dataset for testing our hypotheses is one-word-titles or short titles (ShortT) \cite{Kellert}. One-word titles (henceforth: short titles or ShortT) contain only the target word and no other word can influence the polarity of the target word, e.g. ``Excellent!''. To see how much the linguistic context of other words affects the polarity variation of our target words, we also use datasets with longer titles that we divide into titles without negation words like ``no'' ‘not’, ``nunca'' ‘never’, ``nada'' ‘nothing’ (PosT) and titles with negation words (NegT) (see\cite{kellert2023experimenting} for the list of negation words). 

We end up with three datasets in total (see 6).

\MyQuote
{\textbf{Main dataset} = one-word titles (ShortT), 

\textbf{additional datasets for comparison} = 

titles without negation (PosT), 

titles with negation (NegT)
}

\subsection*{Methods for testing Hypotheses}

Part of the methodology we use in this paper is based on our preliminary study in which we have shown that one-word titles have the smallest ambiguity in comparison to other datasets \cite{Kellert}. For testing our Hypotheses and performing statistical calculations, we first extract words from the datasets in (6) using a conventional lemmatizer and POS-tagger of Spanish from Stanza \cite{qi-etal-2020-stanza}, as we did in our preliminary study \cite{Kellert}. We obtain the following numbers of lemmas for each of our datasets:

\MyQuote
{\textbf{Nr. of lemmas in PosT} = 20,827, 

\textbf{Nr. of lemmas in ShortT} = 3,241, 

\textbf{Nr. of lemmas in NegT} = 2,319 
}

We create dictionary entries for each word that contain various information such as the Standard Deviation (sdv) and Variance of word polarities per word (e.g. sdv/variance ([5, 5, 4, 5, 4, 5, 5,…..])) as shown in (8). The word dictionaries are available for use under § Supporting Information (see, 
\hyperref[S1_File]{S1}, \hyperref[S2_File]{S2}, \hyperref[S3_File]{S3})

as well as on GitHub (\url{https://github.com/olga-kel/LexPol}):

\MyQuote
{\textbf{Word statistics}:

\{'lindo' (engl. ‘cute’): \{'sdv': 0.692, 'variance': 0.484, 'mean': 4.347, 'median': 4, 'frequency': 2979, 'POS': 'ADJ'\}\}}

\begin{figure}[!h]
\caption{{\bf Demonstration of a possible polarity distribution of two word groups (e.g., adjectives vs. nouns)}}
\includegraphics[]{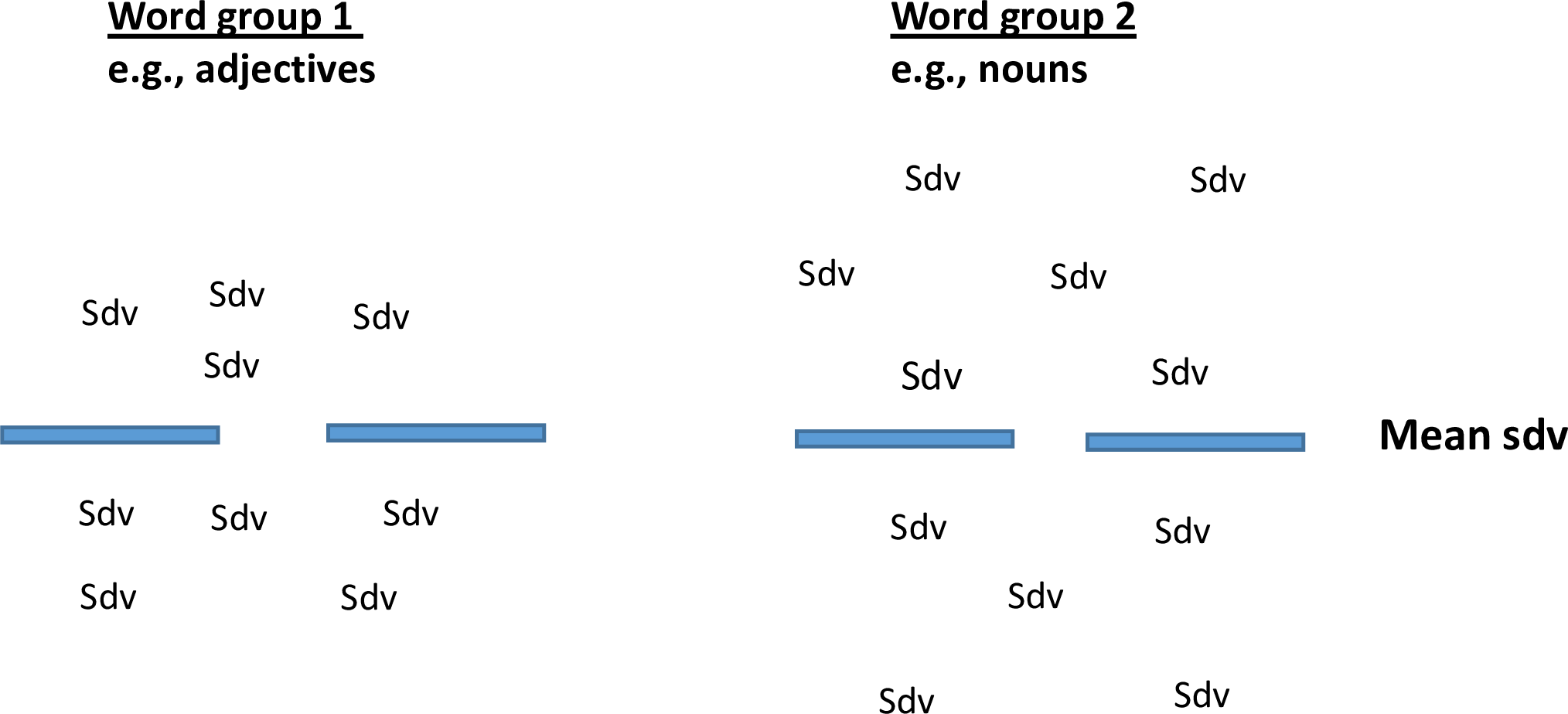}
\label{fig2}
\end{figure}

Sdv characterizes the amount of variation or dispersion of the polarity set per each word. Variance is the expectation of the squared difference of each data point from the mean. Sdv is the square root of the variance. In this sense, both sdv and variance can capture the polarity ambiguity of a word. The higher the sdv or variance of a word, the higher the polarity ambiguity of this word \cite{Kellert}. To characterize the ambiguity of different word groups (see Hypotheses in (5)), we measure the equality of the average of sdv per word group, for which we use Student’s t-test \cite{student1908probable}. The underlying idea behind using the t-test for testing our hypotheses is the assumption that word groups such as adjectives and nouns that are associated with clear polarity values will have a similar polarity distribution or a similar dispersion of the polarity set. For instance, the adjective ``delicious'' and the noun ``deliciousness'' will probably be associated almost only with positive human judgments with a small rate of deviations. In case the word groups have a different amount of polarity ambiguity, they will show different polarity distributions, say the word ``delicious'' and the noun ``restaurant''. The judgments for the latter word might show a full range of polarity variation from positive to negative without any bias towards one polarity. The polarity variation is measured by sdv. We test the Null-Hypothesis that the averages of the sdv of our word groups are equal. This idea is illustrated in Fig~\ref{fig2}.

Before performing the statistical test of the Null-Hypothesis, we calculate the mean of sdv per each group to see whether the means between our groups differ and perform the statistical test using t-test function in Python from the scipy.stats library. The output is something like:

\MyQuote
{\textbf{t-test result}:

(statistic=0.33617324893734357, pvalue=0.5692334858602581)}

If the p-value of the test is less than .05, we reject the Null-Hypothesis according to which the difference between the means of the sdvs of the tested word groups is not statistically significant. 
This means for testing Hypothesis 1 that the mean of sdv of the POS groups we consider, namely adjectives, nouns, adverbs and verbs, will be assumed to be equal in short titles if the Null-Hypothesis is not rejected. If the Null-Hypothesis is rejected, Hypothesis 1 is true or the mean difference of the considered word groups is statistically significant. We focus on POS tags corresponding to content words in Stanza \cite{qi-etal-2020-stanza} [ADJ, VERB, NOUN, ADV]. The reason why we consider only POS groups of content words is that other POS groups like conjunctions or auxiliaries (e.g., ``and'', ``have'') belong to functional POS groups, which rarely encode polarity or subjectivity according to the literature and sentiment dictionaries \cite{lasersohn2005context, kennedy2013two, umbach2020evaluative, brooke-etal-2009-cross} and they are rarely used in one-word titles, which is the main dataset of this study. We tested the assumption that functional words and content words will substantially differ in ambiguity in a pre-test, expecting that the Null-Hypothesis would not be true, that is, the mean of sdv between functional and content words would be equal. Our expectation was confirmed. Functional words and content words show significant differences in the mean of sdv according to the t-test as shown in (10)b. and the functional words have a higher mean of sdv than content words as shown in (10)a.

\MyQuote
{
\textbf{a. Mean of sdv between functional words vs. content words}: 0.96 vs 0.77, 

\textbf{b. Student’s t-test statistic}: -4.47 p=8.0e-06 ($<$\textbf{0.05})
}

The mean of sdv of functional words can be seen as an orientation for the evaluation of content words. Those content words that are close to the mean of functional words can be considered as very ambiguous and those content words that are distant from the mean of functional words are less ambiguous or non-ambiguous. 

Capturing the semantic class of words (Hypothesis 2) is not a trivial task. For testing Hypothesis 2, we created two wordlists from short titles. These wordlists represent the most ambiguous words and the least ambiguous words from short titles according to our measurements, which we introduce below. We assume that the percentage of PTPs or sentiment words and non-PTPs or non-sentiment words will considerably differ in the two wordlists, because PTPs or sentiment words are used to predict subjectivity and/or polarity \cite{brooke-etal-2009-cross} and thus should be ideally non-ambiguous. 

The first wordlist contains words with low sdv (sdv $<$ 0.4) and the second wordlist contains words with high sdv (sdv $>$ 0.8). The reason why we picked these thresholds and not other thresholds is that they fulfill two conditions: they represent two extremes (low ambiguity or low sdv and high ambiguity or high sdv) and at the same time they provide enough words for both word lists. Lower sdv would produce fewer words in the word list representing lower ambiguity and higher sdv would produce more words but less extreme difference with the other word list. The chosen thresholds are thus the optimal trade-of between the two requirements.     
We use three different methods to test Hypothesis 2 or to classify words in our wordlists as PTPs or non-PTPs. Our first method is a manual annotation (Method 1). We manually annotate words with low sdv (sdv $<$ 0.4) and words with high sdv (sdv $>$ 0.8) according to their semantic class as evaluative words or PTPs such as ``tasty'', ``fun'', etc. We follow the definition in the linguistic literature for PTPs \cite{lasersohn2005context,kennedy2013two,umbach2020evaluative,kennedy2022familiarity}. The most important property of PTPs is that they characterize subjective tastes or evaluations of a person or judge and do not represent objectively verifiable facts (compare ``Chocolate is tasty'' vs. ``The earth is round'') (see § Supporting Information, \hyperref[S4_File]{S4}). 
However, manual annotations can be subjective as they depend on the annotator’s perspective, world- and fact knowledge \cite{DBLP:journals/corr/abs-2109-04270}. For this reason, we used two other automatic methods for the classification of PTPs and non-PTPs or testing Hypothesis 2. We used the sentiment dictionary SO-CAL \cite{brooke-etal-2009-cross} to annotate sentiment words in our word lists (Method 2). SO-CAL words and sentiment words in general are usually chosen to predict polarity and thus should ideally function as PTPs \cite{brooke-etal-2009-cross}. We assume that the two wordlists will differ with respect to the amount of sentiment words and non-sentiment words (see Method 1). We should note that even though sentiment dictionaries like SO-CAL are not perfect as they might also contain ambiguous words (see \cite{kellert2023experimenting} on this point), they can still be considered as a good source for classifying words as sentiment words or non-sentiment words due to their usefulness in Sentiment Analysis \cite{kellert2023experimenting}.   
The third automatic method we used was to compare the number of adjectives and other word types such as nouns in the two wordlists (Method 3). The authors in \cite{hill-korhonen-2014-learning} have shown that words expressing subjectivity correlate with adjectives. Our assumption is thus that the two wordlists will differ with respect to their distribution of adjectives and other word types such as nouns. We assume that words with PTPs will contain more adjectives than nouns.

For testing Hypothesis 3, we extract words that have polarity 5, polarity 1 and polarity 3 as the most frequent polarity. Recall that according to Hypothesis 3 neutral words or words with polarity 3 will be more ambiguous than words with extreme polarities 1 and 5. We use The Counter's most common(1) method in Python. For example, the most frequent polarity of the word ``excelente'', [5, 5, 4, 5, 4, 5, 5, 5, 5, 5, 5] is 5. We then test the equality of mean of sdv using Student’s t-test on words that have the extreme polarity 5 and the non-extreme polarity 3 as the most common polarity. We run the same test for words with extreme polarity 1 and non-extreme polarity 3 as the most common polarity, assuming that the mean of sdv will be the same under the Null Hypothesis. If it is not the same, our Hypothesis 3 is confirmed.
In contrast to our preliminary study \cite{Kellert}, we also test the word groups of our interest for the equality of variance of sdv using Levene’s test, which is a statistical test for the equality of group variances \cite{Levene}. Contrary to Student’s t-test, which we use here to compare the level of ambiguity between two word groups (represented by the mean of the sdv), Levene’s test compares the “uniformity of word ambiguity” (represented by the variance of the sdv). This provides us information about whether words from a word group tend to be equally or uniformly ambiguous (low variance of the sdv, regardless of whether its mean is high or low), or, on the contrary, whether the group contains highly ambiguous and non-ambiguous words (high variance of the sdv). In other words, we assume that a lower variance of sdv of a word group x tells us that the level of ambiguity of this word group is very much the same among the words belonging to the word group x. Note, however, that said level of ambiguity could be either high or low: the variance of sdv of the word group x does not give us information about its mean of sdv, hence, both metrics (and their corresponding statistical texts) have orthogonal goals. Fig~\ref{fig2} can be taken as a visual example of data points where we would expect the null hypothesis of the t-test to not be rejected (because sdv means coincide, i.e., both word groups are equally ambiguous on average) whereas the null hypothesis of Levene's test could be rejected (because the sdv distribution is more scattered for one group than the other).
We applied the test on the word groups of content words and functional words. Our assumption is that the mean of sdv of functional words will be higher than that of content words due to their ambiguity \cite{Kellert}, but the variance of sdv of functional words will be lower than the variance of sdv of content words, because functional words are always ambiguous (they carry no polarity by themselves, so polarity in texts containing them can vary greatly depending on context), whereas content words can be either ambiguous like (2) or unambiguous like (1). The results of this pre-test in (11), together with the one previously shown in (10) (repeated here in 12) show that our assumptions are confirmed. Functional words have a lower variance of sdv than content words, but a higher mean of sdv than content words:

\MyQuote
{\textbf{Variance of sdv between functional words (fw) vs. content words (cw)}: 0.09(cw) vs 0.07(fw), 

\textbf{Levene’s test statistic}: 9.21 p=0.0002 ($<$\textbf{0.05})

}

\MyQuote
{\textbf{a. Mean of sdv between functional words vs. content words}: 0.96 vs 0.77, 

\textbf{b. Student’s t-test statistic}: -4.47 p=8.0e-06 ($<$\textbf{0.05})

}

In short, Student’s t-test provides us the results on equality of the means of sdv which tells us about the level of ambiguity of a word group, whereas Levene’s test provides us the results on the equality of variance of sdv which tells us about the uniformity of the ambiguity of a word group. We present the results on the t-test and the results on Levene’s test in §Results and Discussion.



\section*{Results and Discussion}

\subsection*{Hypotheses testing}
Hypothesis 1 is confirmed as the POS influences the polarity variation in short titles (see Table~\ref{table2}). The mean sdv between adjectives and nouns is not equal in short titles and across other datasets (PosT and NegT). As expected, the mean sdv of adjectives is lower than the mean sdv of nouns suggesting that adjectives are less ambiguous than nouns in short titles. The difference in mean of sdv between adjectives and verbs as well as between adjectives and adverbs is not significant in short titles. This suggests that verbs, adverbs and adjectives are very similar in their ambiguity. However, the difference in mean of sdv between adjectives, verbs, and adverbs is significant across longer titles (PosT and NegT), suggesting that the textual context or the length of the sentence influences the polarity variation in such a way to induce a significant difference in polarity variation between adverbs, adjectives, and verbs. Adjectives have a lower mean of sdv in longer titles suggesting that their ambiguity remains low even in the presence of textual context and sentence length variation. This is an important result for Sentiment Analysis and Sentiment Dictionaries as it emphasizes the importance of adjectives in polarity classification (see §Related Work and §Conclusion). However, it is important to note that the effect size of the ambiguity between adjectives and nouns as measured by the mean difference in various datasets (ShortT, PosT, and NegT) is much smaller than the effect size of the ambiguity between content and functional words (mean 0.06 among all three datasets vs. 0.19). This shows that the discrepancy in polarity ambiguity is much higher between functional and content words than between different POS classes of content words.

\begin{table}[!ht]
\centering
\caption{
{\bf Testing Hypothesis 1}}
\begin{tabular}{|p{35pt}|p{65pt}|p{65pt}|p{65pt}|}
\hline
&
{\bf Adj vs. Verbs}&
{\bf Adj vs. Nouns}&
{\bf Adj vs. Adv}\\
\thickhline

\multirow{ 4}{*}{ShortT} & 0.57 (verbs) & 0.56 (nouns) & 0.43 (adv) \\
 & 0.49 (adj) & 0.49 (adj) & 0.49 (adj) \\ 
 & stat=1.28 & stat=3.56 & stat=0.50 \\
 & pval=0.19 & pval=0.0004 & pval=0.61 \\ \hline
\multirow{ 4}{*}{PosT} & 0.85 (verbs) & 0.76 (nouns) & 0.86 (adv) \\
 & 0.73 (adj) & 0.73 (adj) & 0.73 (adj) \\ 
 & stat=5.8 & stat=2.25 & stat=4.25 \\
 & pval=8.6e -9 & pval=0.02 & pval=2.54 e-5 \\ \hline
\multirow{ 4}{*}{NegT} & 1.12 (verbs) & 1.08 (nouns) & 1.14 (adv) \\
 & 0.99 (adj) & 0.99 (adj) & 0.99 (adj) \\ 
 & stat=3.051 & stat=3.057 & stat=3.44 \\
 & pval=0.002 & pval=0.002 & pval=0.0007 \\ \hline
\end{tabular}
\begin{flushleft} Results of Student’s t-test on equality of the mean of sdv for words with different POS across datasets. Each column shows results for different datasets (ShortT, PosT, NegT) with respect to sdv difference between two POS, the statistics and the pvalue.
\end{flushleft}
\label{table2}

\end{table}

Table~\ref{table3} shows that Hypothesis 2 is confirmed using three different independent methods. The amount of PTPs (measured in percent) is higher than the amount of non-PTPs in the word list with lower sdv, whereas the opposite is the case for the wordlist with higher sdv according to method 1 and method 2. The third method does not measure PTPs directly, but the POS, which should correspond to PTPs \cite{hill-korhonen-2014-learning}. The results from the third method confirm that lower ambiguity words contain more adjectives than nouns, whereas higher ambiguity words contain more nouns than adjectives, confirming thus that lower ambiguity words contain more PTPs than higher ambiguity words.

\begin{table}[!ht]
\centering
\caption{
{\bf Testing Hypothesis 2}}
\begin{tabular}{|p{55pt}|p{120pt}|p{120pt}|}
\hline
 & Word freq \textgreater5 and sdv!=0 and \textgreater 0.8 (= high ambiguity) Total=219 &
Word freq \textgreater 5 and sdv!=0 and $<$ 0.4 (=low ambiguity) Total=111\\
\hline
Method 1 = Manual annotation of PTPs & PTSs: 15\% (31/219) vs. non-PTPs: 85\% (183/219)&
PTPs: 71\% (79/111) vs. non-PTPs 29\% (32/111)\\

\hline
Method 2 = SO-CAL Match &  PTPs: 31\% (67/219) vs. non-PTPs: 69\% (152/219)&
PTPs: 54\% (60/111) vs. non-PTPs 46\% (51/111)
\\
\hline
Method 3 = POS Match & Nouns (65\%) vs. Adj.(27\%)
Counter({ 'NOUN': 42, 'ADJ': 61, 'VERB': 8, 'PUNCT': 4, 'INTJ': 2, 'ADV': 1, 'PRON': 1})& Nouns (41\%) vs. Adj.(56\%) Counter({'ADJ': 59, 'NOUN': 46, 'ADV': 2, 'VERB': 2, 'NUM': 1, 'INTJ': 1})

 \\
\hline

\end{tabular}
\begin{flushleft} Results of the relative frequency of PTPs and related POS of PTPs. Each column shows results for different ambiguity levels (high ambiguity and low ambiguity) per method.
\end{flushleft}
\label{table3}
\end{table}

Table~\ref{table4} shows that Hypothesis 3 is confirmed for short titles and all other datasets. The mean sdv of middle polarity words is higher than the mean sdv of extreme polarity words (Polarity 1 and 5) in short titles. However, the mean of sdv of middle polarity words is lower than that of words with extreme polarity words across positive and negative titles suggesting that middle polarity words are less ambiguous than words with extreme polarity in longer titles. 

\begin{table}[!ht]
\centering
\caption{
{\bf Testing Hypothesis 3}}
\begin{tabular}{|p{35pt}|p{65pt}|p{65pt}|p{65pt}|}
\hline
&
{\bf Pol 5 and 1}&
{\bf Pol 3 and 1}&
{\bf Pol 5 and 3}\\
\thickhline

\multirow{ 4}{*}{ShortT} & 0.31 (Pol 1) & 0.31 (Pol 1) & 0.48 (Pol 5) \\
 & 0.48 (Pol 5) & 0.70 (Pol 3) & 0.70 (Pol 3) \\ 
 & stat=2.35 & stat=-5.30 & stat=-4.45 \\
 & pval=0.01 & pval=2.4 e-6 & pval=1.74 e-5 \\ \hline
\multirow{ 4}{*}{PosT} & 1.16 (Pol 1) & 1.16 (Pol 1) & 0.71 (Pol 5) \\
 & 0.71 (Pol 5) & 0.96 (Pol 3) & 0.96 (Pol 3) \\ 
 & stat=12.89 & stat=5.80 & stat=-11.51 \\
 & pval=3.16 e -37 & pval=1.26 e -8 & pval=3.35 e-30 \\ \hline
\multirow{ 4}{*}{NegT} & 1.33 (Pol 1) & 1.33 (Pol 1) & 1.10 (Pol 5) \\
 & 1.10 (Pol 5) & 1.00 (Pol 3) & 1.00 (Pol 3) \\ 
 & stat=4.29 & stat=9.21 & stat=4.40 \\
 & pval=2.20 e-5 & pval=2.39 e-18 & pval=1.21 e-5 \\ \hline
\end{tabular}
\begin{flushleft} Results of Student’s t-test on mean of sdv equality for words with different polarity category across datasets. Each column is showing results for different datasets (ShortT, PosT, NegT) with respect to variance difference, the statistics and pvalue.
\end{flushleft}
\label{table4}
\end{table}

\begin{figure}[!h]
\caption{{\bf Number of POS instances (ADJ, NOUN, VERB, ADV, FUNCTIONAL) per polarity in NegT}}
\includegraphics[]{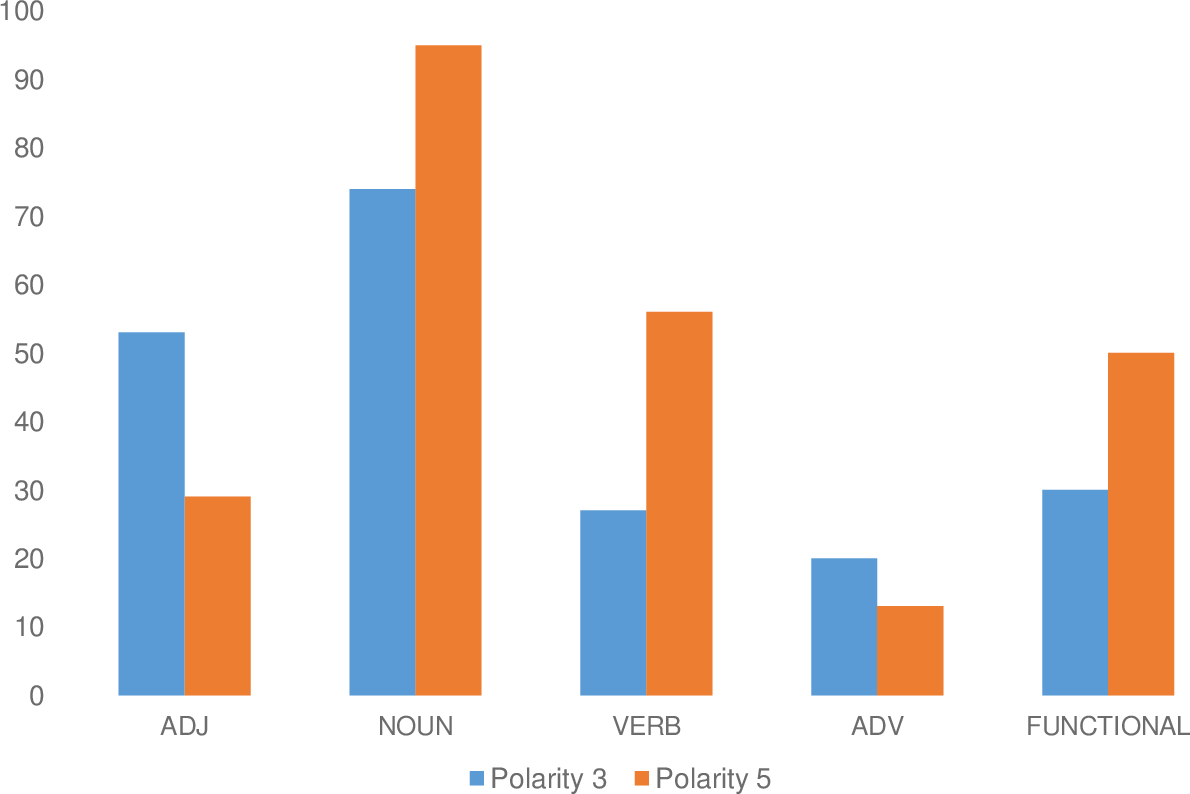}
\label{fig3}
\end{figure}

In order to understand why middle polarity words are less ambiguous than words with polarity 5 in NegT (see Table~\ref{table4}, last row), we have inspected negative titles and have found that titles from NegT with polarity 5 often contain recommendations with verbs like ``No puedes perderselo!'' ``You can’t miss it!'', whereas titles from NegT with polarity 3 often contain negation of adjectives like ``No tan bonito'' ‘Not so nice’ or ``Bonito, pero no mucho'' ‘Nice, but not so much.’. Given the results from our Hypothesis 1 that have shown statistical differences between adjectives and verbs in NegT, we assumed that the POS could be the triggering factor of the difference in mean of sdv between middle polarity words and words with positive polarity in NegT. This assumption is confirmed in Fig~\ref{fig3} showing that adjectives are more frequent than verbs in the case of words with polarity 3 but not in the case of of words with polarity 5 (see, § Supporting Information, \hyperref[S5_File]{S5}). 
In short, POS is the trigger of the difference in mean of sdv between words with polarity 3 and 5 taken from NegT.

Note that the effect size of the ambiguity between words of middle polarity (Pol 3) and words of extreme polarities (Pol 1 and Pol 5) as measured by the mean difference in ShortT is bigger than the effect size of the ambiguity between content and functional words (mean 0.26 vs. 0.19). This shows that the discrepancy in polarity ambiguity is higher between different polarity categories (middle and extreme polarities) than between functional and content words. If we compare the mean differences between polarity categories with the mean differences between POS classes in ShortT, we observe that the effect size of the ambiguity between polarity classes is much higher than between POS classes such as nouns and adjectives (0.07 vs. 0.26) (compare Table~\ref{table2} first row and Table~\ref{table4} first row). The differences in effect size of POS classes, polarity classes and functional/content word classes thus show that the effect size of polarity categories is the highest among other classes. This is a new observation that we discuss in §Conclusion.

We summarize our results for testing our Hypotheses in Table~\ref{table5}.

\begin{table}[!ht]
\centering
\caption{
\textbf{Overall results of our Hypotheses. 
}}
\begin{tabular}{|p{35pt}|p{55pt}|p{55pt}|p{110pt}|}
\hline
 &
 \begin{tabular}{p{50pt}}
H1\\
POS influences ambiguity\\
\end{tabular}
&
\begin{tabular}{p{52pt}}
H2\\
Semantic class influences ambiguity\\
\end{tabular}
&
\begin{tabular}{p{103pt}}
H3\\
Polarity category influences ambiguity\\
\end{tabular}
\\
\hline
Short tiles &
\checkmark between adjectives and nouns
&
\checkmark
&
\begin{tabular}{p{104pt}}
\checkmark all categories (1, 3, 5)\\
(\# the mean of sdv of Pol 3 is higher than the mean of sdv of Pol. 5 and 1 in short titles as expected)
\end{tabular}\\
\hline

Longer tiles &
\checkmark for all POS
&
Not tested
&
\begin{tabular}{p{104pt}}
\checkmark all categories (1, 3, 5)\\
(\# the mean of sdv of Pol 3 is smaller than the mean of sdv of Pol. 5 in negative titles due to POS)
\end{tabular}\\
\hline
\end{tabular}
\begin{flushleft} Overall results of our hypotheses.
\end{flushleft}
\label{table5}
\end{table}

\subsection*{Levene’s Test on uniformity of ambiguity}
To complement our results, we now apply Levene's test which, as explained above, can be used to compare word groups in terms of uniformity of polarity ambiguity. Table~\ref{table6} shows that adjectives have statistically smaller variance of sdv than nouns in short titles and across other datasets (e.g. negative titles), which suggests that adjectives are uniformly unambiguous or that their low ambiguity level does not vary much from one word to another word. This is different with nouns. The variance of sdv is higher for nouns than for adjectives, which suggests that their ambiguity level changes much more relative to adjectives. We will explain this effect given that nouns can be used as topic words such as ``servicio'' ‘service’ or as sentiment words ``excelencia'' ‘excellency’ and thus change from highly ambiguous to non-ambiguous (see §Conclusion).

\begin{table}[!ht]
\centering
\caption{
{\bf Results on variance differences of words with different POS across datasets}}
\begin{tabular}{|p{35pt}|p{65pt}|p{65pt}|p{65pt}|}
\hline
&
{\bf Adj vs. Verbs}&
{\bf Adj vs. Nouns}&
{\bf Adj vs. Adv}\\
\thickhline

\multirow{ 4}{*}{ShortT} & 0.09 (verbs) & 0.10 (nouns) & 0.16 (adv) \\
 & 0.06 (adj) & 0.06 (adj) & 0.06 (adj) \\ 
 & stat=3.85 & stat=14.89 & stat=1.71 \\
 & pval=0.05 & pval=0.0001 & pval=0.19 \\ \hline
\multirow{ 4}{*}{PosT} & 0.08 (verbs) & 0.10 (nouns) & 0.08 (adv) \\
 & 0.08 (adj) & 0.08 (adj) & 0.08 (adj) \\ 
 & stat=2.73 & stat=6.43 & stat=0.08 \\
 & pval=0.09 & pval=0.01 & pval=0.77 \\ \hline
\multirow{ 4}{*}{NegT} & 0.13 (verbs) & 0.07 (nouns) & 0.08 (adv) \\
 & 0.05 (adj) & 0.05 (adj) & 0.05 (adj) \\ 
 & stat=20.9 & stat=4.83 & stat=3.82 \\
 & pval=8.12 e-6 & pval=0.03 & pval=0.05 \\ \hline
\end{tabular}
\begin{flushleft} Results on variance differences of words with different POS across datasets.
\end{flushleft}
\label{table6}
\end{table}

Table~\ref{table7} shows that the uniformity of ambiguity across words with different polarity is the same in short titles, meaning that their ambiguity level does not change much. Words with middle and extreme polarities are uniformly ambiguous in short titles. However, they don’t have a uniform ambiguity level in datasets with longer titles. In positive titles, middle polarity words show a more uniform ambiguity than words with extreme polarities 1 and 5 suggesting that the polarity variation of middle polarity words is much more robust than the polarity variation of extreme polarity words.

\begin{table}[!ht]
\centering
\caption{
{\bf Results on variance differences of words with different polarities across datasets}}
\begin{tabular}{|p{35pt}|p{65pt}|p{65pt}|p{65pt}|}
\hline
&
{\bf Pol 5 and 1}&
{\bf Pol 3 and 1}&
{\bf Pol 5 and 3}\\
\thickhline

\multirow{ 4}{*}{ShortT} & 0.05 (Pol 1) & 0.05 (Pol 1) & 0.08 (Pol 5) \\
 & 0.08 (Pol 5) & 0.06 (Pol 3) & 0.06 (Pol 3) \\ 
 & stat=0.88 & stat=0.46 & stat=0.21 \\
 & pval=0.34 & pval=0.05 & pval=0.64 \\ \hline
\multirow{ 4}{*}{PosT} & 0.18 (Pol 1) & 0.18 (Pol 1) & 0.12 (Pol 5) \\
 & 0.12 (Pol 5) & 0.05 (Pol 3) & 0.05 (Pol 3) \\ 
 & stat=14.39 & stat=70.67 & stat=49.06 \\
 & pval=0.00 & pval=7.4 e -16 & pval=2.91 e-12 \\ \hline
\multirow{ 4}{*}{NegT} & 0.12 (Pol 1) & 0.12 (Pol 1) & 0.05 (Pol 3) \\
 & 0.13 (Pol 5) & 0.05 (Pol 3) & 0.13 (Pol 5) \\ 
 & stat=0.10 & stat=21.86 & stat=58.20 \\
 & pval=0.74 & pval=4.12 e-6 & pval=8.4 e-14 \\ \hline
\end{tabular}
\begin{flushleft} Results on variance differences of words with different polarities across datasets.
\end{flushleft}
\label{table7}
\end{table}

In short, we have shown that adjectives are uniformly unambiguous across all datasets and that words with polarity categories 1, 3 and 5 have a uniform ambiguity level in short titles.


\section*{Conclusion}


Hypothesis 1 has been confirmed as expected. POS matters for polarity variation as adjectives and nouns show significant differences in the mean of sdv. The mean of sdv is higher for nouns than for adjectives. Nouns can be very ambiguous if they are used as topic words (e.g., ``As to the service, I like it'' or ``As to the service, I don’t like it.'') and not for characterization of a target, hence the stronger ambiguity of nouns than adjectives in short titles. Interestingly, the same effect holds for other datasets as well (positive and negative titles), suggesting that the influence of context words does not change the difference in ambiguity between adjectives and nouns. In addition, we have shown that adjectives have a lower variance than nouns suggesting that they are uniformly unambiguous, whereas nouns are less uniform showing that some nouns are more ambiguous than others. The latter observation is coherent with nouns having different functions in titles (topic words and sentiment words).

Verbs and adverbs did not show any difference in ambiguity from adjectives in short titles suggesting that these POS groups are well chosen to characterize the target of evaluation in short titles. This result can be used to enhance sentiment dictionaries like SO-CAL \cite{brooke-etal-2009-cross} by adverbs and verbs from short titles. Our results showing that verbs, adverbs, and adjectives have the same ambiguity level in short titles contradict previous results in the NLP literature according to which adjectives and verbs belong to the most ambiguous word group \cite{wu-jin-2010-semeval, Xia}. We assume that this discrepancy is due to the lack of statistical methods and the lack of control for sentence length in previous studies as well as due to the methodological differences chosen for testing POS ambiguity. This conclusion is confirmed by the choice of the most ambiguous adjectives in previous research on polarity ambiguity. The authors in \cite{wu-jin-2010-semeval, Xia} provide examples of ambiguous adjectives like long, short, few, big, huge, small, tall, etc, which belong to a particular adjective group in linguistic research, namely to adjectives that can be modified by measure phrases like ``180 cm tall'' \cite{kennedy2013two}. This adjective group is known for being ambiguous between subjective and objective interpretation \cite{kennedy2013two}. The sentence ``He is 180 cm. tall'' is an objective statement that does not represent a personal opinion as in ``I find him tall'' \cite{kennedy2013two}. Moreover, depending on the linguistic neighbor, the adjectives ``big'' and ``small'' considered ambiguous words in previous literature can be interpreted positively or negatively, e.g. ``big problem'' vs. ``big talent''. As our experiments consider words from TripAdvisor reviews, especially from one-word titles, and not all adjectives in general that might exist in a language, this methodological focus might explain the discrepancy between our results and results in previous studies. Short titles contain more qualitative adjectives than adjectives combinable with measure phrases such as ``tall''. In this sense, our study focuses on the distinction between POS classes in a particular context such as one-word titles context. In this context, adjectives are less ambiguous than nouns and as ambiguous as verbs and adverbs. 

Our results also show that, in longer sentences, verbs and adjectives show differences in ambiguity suggesting that verbs are more ambiguous than adjectives in longer sentences. We have shown that this effect correlates with polarity differences of words in negative titles (Fig~\ref{fig3}). According to our data in Table~\ref{table2}, adverbs and adjectives do not show any difference in the mean of sdv in short titles. We can explain this effect, given that adverbs in short titles have content information like ``great'' or ``super'' and encode polarity lexically. However, there is a difference between adverbs and adjectives in longer titles, which needs to be explained in the future. In short, the results from testing Hypothesis 1 have provided new information about the differences of POS with respect to their ambiguity level and ambiguity uniformity in short titles and other datasets in Spanish that has been missing in Linguistics and NLP (§Related Work).

We have tested Hypothesis 2 according to which evaluative predicates or PTPs encode subjectivity and polarity lexically showing that evaluative predicates or PTPs are relatively more frequent among words with the lowest ambiguity (sdv $<$ 0.4) than among words with the highest ambiguity (sdv $>$ 0.8) in short titles. Moreover, words with the lowest ambiguity match relatively more often with sentiment words than words with the highest ambiguity and the words with the lowest ambiguity contain more adjectives than nouns. These results confirm the observation in NLP showing that subjective words often correlate with adjectives \cite{hill-korhonen-2014-learning}. Our results provide new empirical support for the importance of PTPs in studies on subjectivity \cite{lasersohn2005context,kennedy2013two,umbach2020evaluative,kennedy2022familiarity}. One interesting result from the study on Hypothesis 2 is that the classification of PTPs and non-PTPs of short titles by sentiment words from SO-CAL dictionary (method 2) produced a less strong difference between PTPs and non-PTPs in short titles than the manual annotation in method 1. For instance, the discrepancy between PTPs and non-PTPs is much higher for words with lower ambiguity in the first than in the second method (see Table~\ref{table3}). This result can be interpreted as a confirmation that what counts as a PTP or as a sentiment word is subjective by itself and it emphasizes the difficulty of defining sentiment words that are inherently subjective. At the same time, the observation that both methods converge on the overall classification and the results for testing Hypothesis 2 shows that there is some agreement among humans on what counts as a PTP or sentiment word. We leave the analysis of matches and mismatches between manual annotations (method 1) and annotations by SO-CAL (method 2) for future research.    

Our results from testing Hypothesis 3 confirm Hypothesis 3 on the basis of short titles. Middle polarity words are indeed more ambiguous than words with extreme polarities confirming the intuition of the difference between (1) and (2) and the literature in §Related Work. This result has practical implications for lexicographers or sentiment dictionary builders. We suggest a revision of existing sentiment words with middle polarity values like ``ok'', ``average'', ``normal'', etc. in sentiment dictionaries and the development of a method for dealing with polarity ambiguous words in sentiment dictionaries \cite{VilaresPSC18}. Our recent results from testing dictionaries in polarity prediction tasks show that considering the polarity ambiguity of sentiment words improves the accuracy of dictionary-based approaches \cite{Imran}.  By testing the variance difference of sdv among words with polarity 1, 3 and 5, we have shown that the uniformity of ambiguity is equal among these words in short titles. This suggests that these word groups are equally uniform in their ambiguity.

However, our results from testing Hypothesis 3 on longer titles produced different results than expected, as the mean of sdv for words with polarity 3 is lower than the mean of sdv for words with polarity 1 and 5 in longer titles. We have explained this effect showing that it correlates with POS (see Fig~\ref{fig3}). However, according to \cite{alvarez2023overview}, all NLP approaches that participated in the Shared Task of Rest-Mex 2023 made bigger errors in predicting in-between polarities like 2, 3, and 4 than predicting polarity 1 and 5 of the Rest-Mex 2023 test dataset (see §Related Work). One possibility to explain the discrepancy between our results on testing Hypothesis 3 on longer titles and the error analysis of \cite{alvarez2023overview} is to assume that the error analysis would change substantially if only the polarity 3 would be compared to polarity 5 and 1 and not the sum of all in-between polarities (including polarity 2 and 4). We suspect that predicting polarity 2 and 4 is indeed harder than predicting polarity 1 and 5. However, predicting polarity 3 might be easier than predicting polarity 5 in negative titles. The error analysis of the polarity classification of the Rest-Mex dataset needs to be analyzed more closely in the future. Another possible explanation of the conflicting results is that the error analysis mentioned by \cite{alvarez2023overview} and our results from testing Hypothesis 3 are not comparable as we focus in our study on words taken from titles, whereas the source of polarity classification in the Rest-Mex shared task is based on entire reviews. In order to test the second explanation, we need to use our methodology on entire reviews to see any difference in polarity variation. This step is reserved for future work. 

The discussion of the differences in effect sizes measured by the mean difference of tested word groups as represented in Table~\ref{table2} and Table~\ref{table4} has shown that the effect size is the highest among polarity categories, compared to those among functional and content word classes and POS groups. This result suggests that the polarity category (middle and extreme polarities) has a much more important influence on ambiguity than the POS classes. To our knowledge, this discovery has not been observed in linguistic research and Sentiment Analysis in NLP yet \cite{lasersohn2005context,kennedy2013two,umbach2020evaluative,kennedy2022familiarity, cui2023survey,vilares2017universal,kellert2023experimenting,hutto2014vader}.  

Our results from this study have contributed to the theoretical research on subjectivity and polarity and they are useful for NLP practitioners who base their analysis on sentiment dictionaries or hybrid approaches for polarity classification tasks in Sentiment Analysis. More precisely, our results can be used for testing how much NLP approaches that use linguistic features such as POS, sentence length, polarity category, content vs. functional word distinction and the semantic class (+/- PPTs) can improve performances. Testing approaches in Sentiment Analysis, especially in the polarity classification tasks enriched with linguistic information, will be done in the future. 

Our work also contributes to the new rising field of analyzing and modeling judgment variation in NLP in order to capture individual preferences and biases of annotators or judgment providers \cite{DBLP:journals/corr/abs-2109-04270}. We hope that our work has contributed to this field as well, by studying the impact of linguistic factors influencing judgment variation.

Some limitations in our study need to be addressed in future research. The present study focuses on Spanish TripAdvisor Reviews that are biased toward positive statements. It would be interesting for future work to confirm whether our findings generalize to other languages and domains or not. We will apply our methodology to other sentiment datasets such as complaints to get more words with negative polarity, which have a low representation in the datasets used in this study. 

To summarize the key findings of the present study and to emphasize their significance in the broader context of sentiment analysis, we have tested the influence of three linguistic variables on the polarity variation of words, which reflects lexical ambiguity, namely the POS (e.g. adjectives and nouns), the semantic class (+/- PTPs) and the polarity category (middle and extreme polarity categories). We found out that all three variables influence the polarity variation or the lexical ambiguity. Our results show that adjectives are less ambiguous than nouns, words with extreme polarities are less ambiguous than words with middle polarities and sentiment words are less ambiguous than other words. The polarity category has a bigger effect than other variables on the ambiguity if we consider the effect size or the mean difference in sdv. These key findings can be implemented for the improvement of polarity prediction tasks in Sentiment Analysis in dictionary-based and hybrid-based approaches.

\section*{Supporting information}


\paragraph*{S1.}
\label{S1_File}
{\bf ShortT} 

\paragraph*{S2.}
\label{S2_File}
{\bf PosT} 

\paragraph*{S3.}
\label{S3_File}
{\bf NegT} 

\paragraph*{S4.}
\label{S4_File}
{\bf Human annotations} 

\paragraph*{S5.}
\label{S5_File}
{\bf NumbersFigure3} 

\section*{Acknowledgments}
We thank Nicholas Hill Matlis for fruitful discussions on the interpretation of Levene’s test results in the domain of lexical ambiguity.


 


%
%
%

\bibliography{refs}






\end{document}